\documentclass[runningheads]{llncs}
\usepackage[T1]{fontenc}
\usepackage{graphicx}

\begin{document}
\title{Deep Learning for Cardiovascular Risk Assessment: Proxy Features from Carotid Sonography as Predictors of Arterial Damage}
\titlerunning{Deep Learning for Cardiovascular Risk Assessment}
%

\author{Christoph Balada\inst{1}\orcidID{0000-0003-0307-7866} \and
Aida Romano-Martinez\inst{2,3} \and
Vincent ten Cate\inst{2,3} \and
Katharina Geschke\inst{2} \and
Jonas Tesarz\inst{2} \and
Paul Claßen\inst{2} \and
Alexander K. Schuster\inst{2} \and
Dativa Tibyampansha\inst{2} \and
Karl-Patrik Kresoja\inst{2} \and
Philipp S. Wild\inst{2,3} \and
Sheraz Ahmed\inst{1} \and
Andreas Dengel\inst{1}\orcidID{0000-0002-6100-8255}}
\authorrunning{C. Balada et al.}
%
\institute{
German Research Center for Artificial Intelligence (DFKI), 67663 Kaiserslautern, Germany \and
University Medical Center of the Johannes Gutenberg-University Mainz, Germany \and
German Center for Cardiovascular Research (DZHK), Germany}

%
\maketitle              

\begin{abstract}
In this study, hypertension is utilized as an indicator of individual vascular damage. 
This damage can be identified through machine learning techniques, providing an early risk marker for potential major cardiovascular events and offering valuable insights into the overall arterial condition of individual patients. 
To this end, the VideoMAE deep learning model, originally developed for video classification, was adapted by finetuning for application in the domain of ultrasound imaging. 
The model was trained and tested using a dataset comprising over 31,000 carotid sonography videos sourced from the Gutenberg Health Study (15,010 participants), one of the largest prospective population health studies. 
This adaptation facilitates the classification of individuals as hypertensive or non-hypertensive ($75.7\% $ validation accuracy), functioning as a proxy for detecting visual arterial damage.
We demonstrate that our machine learning model effectively captures visual features that provide valuable insights into an individual's overall cardiovascular health.

\end{abstract}

\section{INTRODUCTION}

Cardiovascular diseases (CVD) are the leading cause of death globally, accounting for 32\% of all deaths in 2019 \cite{CVDfactSheet}. 
Among the numerous manifestations of CVD, thrombosis — a condition marked by the formation of blood clots within vessels — plays a pivotal role in causing myocardial infarctions (MI) and ischaemic strokes, which are major contributors to morbidity and mortality worldwide. 
The carotid arteries, which supply blood to the brain, are often involved in atherosclerotic disease, leading to narrowing or occlusion that heightens the risk of stroke \cite{stein2008use}. 
Accurate assessment of carotid artery health is therefore critical for early detection and management of CVD and associated complications.

Carotid ultrasonography is a widely used, non-invasive imaging modality for evaluating carotid artery structure and function. 
By visualizing arterial plaques and measuring blood flow, carotid ultrasound plays a central role in identifying individuals at risk of stroke or other thrombotic events \cite{nezu2020usefulness}. 
However, interpreting carotid ultrasound videos requires significant expertise, as it involves analysing dynamic, high-dimensional data for subtle pathological features, such as intima-media thickening and hemodynamic disturbances \cite{abbott2015systematic}.
Furthermore, ultrasound videos contain a multitude of high-frequency textural features that are imperceptible to the human eye and even to experts with considerable expertise \cite{loizou2015atherosclerotic}. 

Recent advancements in artificial intelligence (AI) have transformed medical imaging by facilitating automated, precise, and reproducible analyses of complex datasets. 
AI methodologies are increasingly applied to a range of ultrasound-based tasks, including classification \cite{gan2024wal}, prediction \cite{lin2022applying}, and segmentation \cite{jain2022attention}. 
Nevertheless, the efficacy of AI is inherently reliant upon the accessibility of substantial amounts of labelled data, which, particularly within the medical domain, can prove to be either impractical or prohibitively costly and time-consuming to obtain. 
This phenomenon is also evident in the domain of carotid ultrasonography, which is why artificial intelligence (AI) models that have been trained and evaluated on just a few hundred images have become the standard in many publications \cite{jain2022attention,gan2024wal}.

This study presents a novel machine learning model that analyses carotid ultrasound videos rather than images, thereby facilitating the extraction of more sophisticated features for a range of downstream tasks.
The model utilises 31,019 videos from the Gutenberg Health Study (GHS) \cite{wild2012gutenberg}, a unique large-scale dataset in the domain of prospective and representative population studies. 
In light of our findings, we propose a novel digital biomarker that leverages hypertension as a proxy for assessing individual risk of CVDs. 
Furthermore, we advocate for additional research focusing on AI-driven assessment of ultrasound video data at the individual patient level.

\section{BACKGROUND}
Video-based classification involves assigning a specific class label to sequences of frames within a video, enabling the identification of particular conditions or characteristics. 
In this study, we developed a machine learning pipeline designed to process video imagery obtained from medical ultrasound devices. 
To achieve this, we finetuned the established VideoMAE model \cite{tong2022videomae}, enabling the analysis of carotid artery ultrasound videos for the classification of individuals as hypertensive or non-hypertensive.
VideoMAE, a video-based machine learning model, employs self-supervised pre-training to efficiently extract spatiotemporal features from video data. 
This capability renders it highly suitable for tasks requiring nuanced temporal analysis and aligns well with the specific requirements of our application.
By focusing on the dynamic flow patterns and morphological changes captured in the ultrasound sequences, our approach aligns with the goals of advanced classification systems in medical imaging.

Transfer learning principles, integral to our model, provide substantial benefits by leveraging pre-trained models on large-scale datasets, significantly enhancing efficiency and accuracy when applied to target tasks with limited data. 
In our study, we utilize the GHS one of the largest and most comprehensive ultrasound datasets available, enriched with a diverse array of clinical information. 
This dataset not only supports robust model training but also enables nuanced analysis of video-based medical imaging. 
By harnessing the temporal and spatial insights captured in these ultrasound sequences, we establish a new milestone in large-scale ultrasound analysis.


\section{METHOD}
Carotid artery ultrasound sonography is a key component of contemporary cardiovascular risk assessment. 
Current practices primarily focus on evaluating intima-media thickness (IMT), vascular stiffness, and the presence of plaques or stenosis to estimate an individual’s statistically elevated risk relative to the expected arterial condition for their age \cite{bao2023carotid}. 
However, the use of visual features from ultrasound imaging for individualized risk prediction is not commonly employed.

The objective of this study is to explore novel machine learning methodologies to characterize cardiovascular risk at the individual level, leveraging the actual visual features captured in ultrasound imaging. 
Hypertension, a principal driver of damage in the vessel and a precursor to plaque development \cite{pfisterer2013einfluss}, is used in this study as a proxy-indicator of individual vascular damage. 
Machine learning approaches enable the detection of vessel damage, even when such damage is imperceptible to the human eye in ultrasound images. 
These early indicators may provide valuable insights into future major cardiovascular events and the overall arterial condition of individuals.

To achieve this, we set up a machine learning pipeline, to classify individuals as hypertensive or non-hypertensive. 
This classification serves as a proxy for identifying visible and sub-visible damage in the vessel and surrounding tissue.
In order to ensure a consistently clear distinction, visual damage will be referred to in the following when the classification result of the AI model is addressed. 
An individual who has been classified as hypertensive by the AI model is therefore referred to as an individual with high visual damage.
This serves to ensure clarity between the output of the AI model and a clinical diagnosis of hypertension.

\begin{figure}[!b]
  \centering
  \includegraphics[scale=.6,trim=0cm 0cm 1cm 0cm,clip]{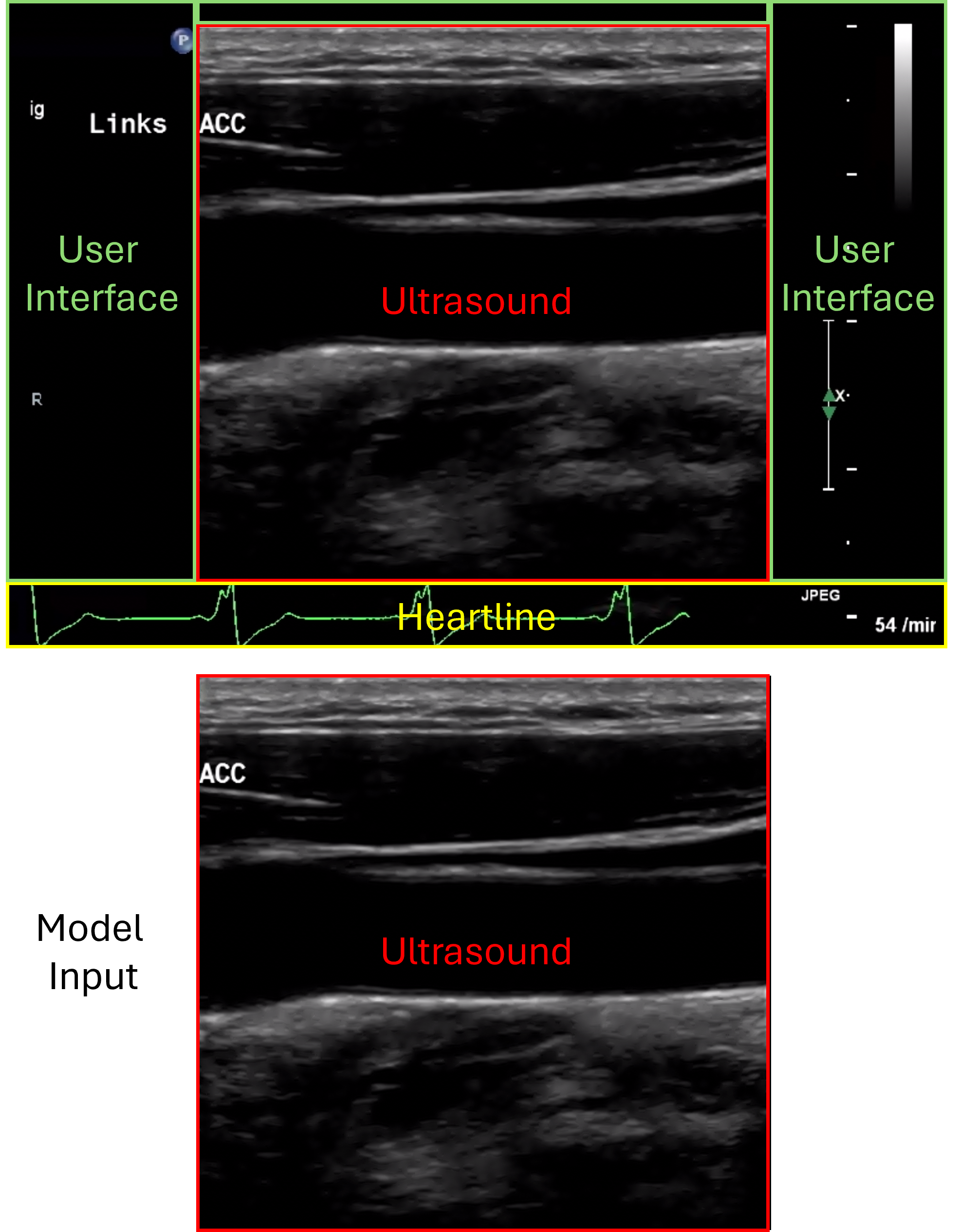}
  \caption{Video before (top) and after (bottom) preprocessing. We remove the user interface and the heartline to avoid biases.}
  \label{fig:preprocessing}
\end{figure}

\subsection{Dataset}
The GHS \cite{wild2012gutenberg} is a large-scale, prospective population-based cohort study initiated in April 2007 by University Medical Center Mainz.
With 15,010 participants, it is among the largest local health studies globally. 
The first phase (2007–2017) involved two comprehensive participant evaluations, while the second phase, currently underway, includes a third evaluation. 
The study focuses on the health status and disease progression within the Rhine-Main region (Germany), with primary emphasis on cardiovascular health. 
Its goal is to identify risk factors and causes of common diseases, contributing to preventive healthcare. 
As part of these comprehensive evaluations, each individual underwent a carotid sonography assessment, where multiple ultrasound videos from different perspectives were collected.
In addition to the ultrasound assessment, individuals were screened for anthropometric characteristics, traditional risk markers, comorbidities, laboratory parameters, and also future incidents. 
The presence of risk factors is determined by a combination of self-reported information, in-house biomarker measurements as well as medication intake. 
Incidence data on cardiovascular events were assessed via structured follow-ups with subsequent validation of endpoints. 
All-cause death was obtained via monthly checks with German registration offices. 
Cardiac death was determined via quarterly review of death certificates using ICD-10 coding.
Additional details regarding the study design can be found in a separate publication \cite{wild2012gutenberg}.

\subsection{Data Preprocessing \& Training}
The raw videos extracted from the DICOM files initially retained elements of the ultrasound device’s user interface, including the heartline displayed on-screen. 
To address this, the user interface and heartline were removed by applying pixel-change thresholding over time. 
Additionally, 45 pixels from the bottom of each frame were cropped to eliminate any remaining portions of the heartline overlay. 
Videos containing Doppler visualizations were excluded by filtering out those exceeding a specified threshold of red or blue pixels in the HSV colour space. 
These steps were implemented to avoid biases or misinterpretations potentially introduced by the presence of the heartline.

Figure \ref{fig:preprocessing} illustrates an example of the data before and after the removal of the user interface. 
From each processed video, multiple clips of 16 frames, representing a duration of $2.1s$, were uniformly sampled. 
The videos were normalized using the mean and standard deviation values estimated during the pre-training on the Kinetics-600 dataset.

During training, data augmentation techniques were applied, including random short-side scaling, random cropping to a resolution of $224 \times 224$, and random horizontal flipping. 
Weighted random sampling was employed to address class imbalances in the dataset. 
Training was conducted over eight epochs, with evaluations performed five times during the training process. 
The final model was selected based on the best-performing evaluation results.

\subsection{Statistical analysis}
To evaluate the alignment of the model’s extracted features with cardiovascular risk, we conducted statistical comparisons of the clinical diagnosis of hypertension (hypertensive vs. non-hypertensive) and the degree of visual impact on the vessel (low vs. high visual damage (VD)). 

To assess the arterial health condition of each individual group, we analyse and compare various clinical parameters across the following categories:
\subsubsection{Laboratory parameters:} troponin I and NT-proBNP.
\subsubsection{Comorbidities:} atrial fibrillation, congestive heart failure, past MI, past stroke, coronary artery disease, and CVD.
\subsubsection{Traditional risk factors:} dyslipidemia, diabetes type 2, positive family history of MI or stroke and SCORE2-Germany  (applicable to individuals without prior CVD or diabetes, aged 40–69 years).
\subsubsection{Carotid sonography:} total plaque count. \\

For each parameter, the distribution of values (represented by quartiles: 25\%, median, 75\%) or the prevalence of specific conditions is systematically compared across groups. Additionally, GHS provides data on future cardiovascular events, including stroke (within a 5-year period), MI (over 5 years), and cardiac death (at 5- and 10-year intervals). 
To evaluate the model’s performance, the proportions of individuals within each of the four groups experiencing these events, as recorded by the GHS, are compared.


\section{RESULTS}
\begin{figure}[!b]
  \centering
  \includegraphics[scale=.5,trim=0cm 0cm 0cm 0cm,clip]{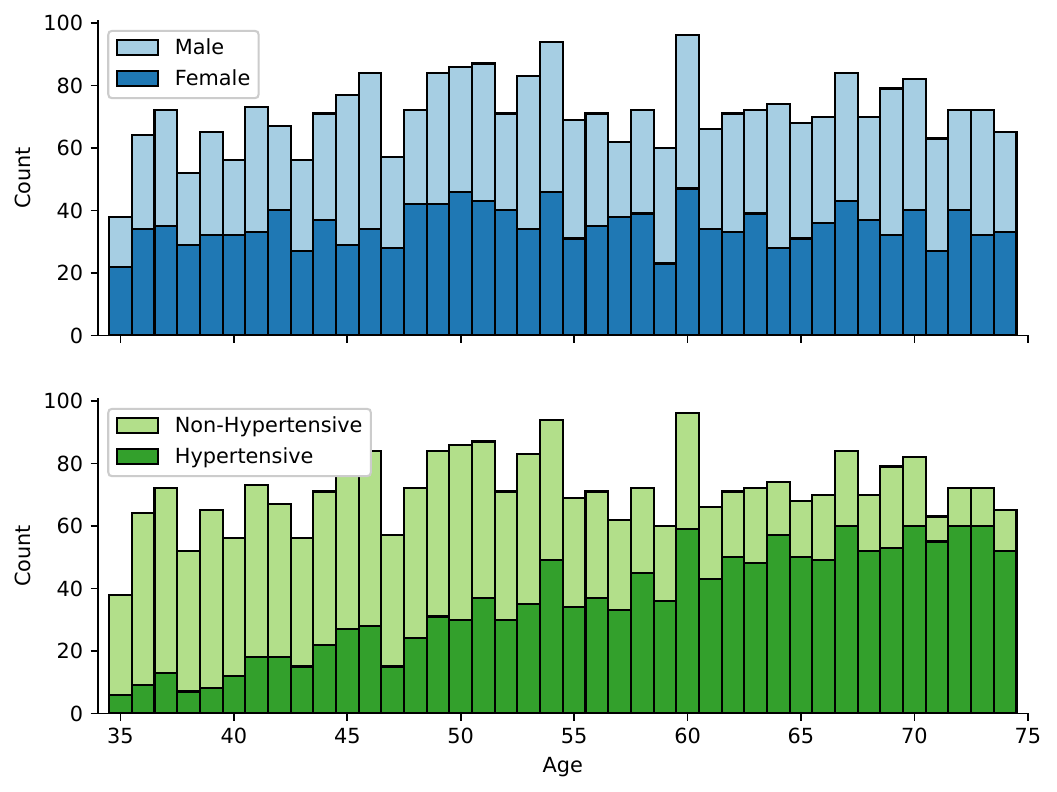}
  \caption{Distribution of individuals' gender and hypertension diagnosis according to age. The mean age of the validation population was found to be $55.5\pm11.2$ years, with $46.9\%$ of the female population and $55.9\%$ of the male population diagnosed with hypertension. The validation dataset revealed a total of $49.2\%$ women.}
  \label{fig:val_data}
\end{figure}

\subsection{Data}
In this study, we utilize ultrasound video data obtained from the GHS initial assessments, comprising 31,019 videos in total. 
These videos are distributed across different anatomical regions, including the left and right common carotid artery (CCA), external carotid artery (ECA), and internal carotid artery (ICA). 
Notably, the majority of the videos (approximately 87\%) are evenly divided between the left CCA and right CCA.
For evaluation purposes, the dataset was partitioned at the level of individuals, with 80\% of the data (11,398 individuals) allocated to the training set and the remaining 20\% (2,847 individuals) assigned to the validation set.
This method ensures that no data leakage occurs between training and validation sets, thereby preventing the introduction of selection bias that could arise from manual splitting.
Figure \ref{fig:val_data} presents the validation dataset, which was randomly sampled from the GHS baseline dataset.
To address class imbalances, balanced accuracy was employed as the criterion for selecting the optimal model.

\subsection{Performance}
\begin{table}[!b]
    \centering
    \begin{tabular}{c|cc}
        Model & $ACC_{val}\uparrow$ & $bACC_{val}\uparrow$ \\ \hline
        VideoMAE & \textbf{75.7\%} & \textbf{72.8\%} \\ 
        ViViT & 68.7\%  & 37.5\% \\
        TimeSformer & 65.9\% & 31.7\% \\
    \end{tabular}
    \caption{Accuracy and balanced accuracy on the validation set for different backbone architectures}
    \label{tab:performance}
\end{table}
\begin{figure}[!b]
  \centering
  \includegraphics[scale=.6,trim=0cm 0cm 2cm 0cm,clip]{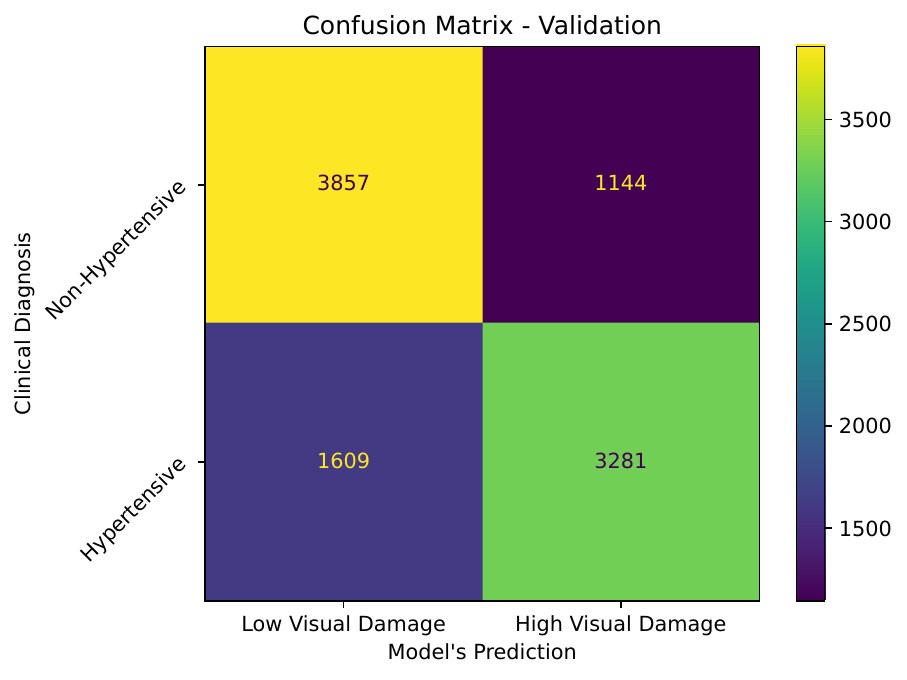}
  \caption{Confusion matrix for the trained model on validation data. Counts are given at sample level. During an evaluation, our pipeline samples uniformly 9891 samples from 6231 unique videos from 2847 individuals in the validation dataset.}
  \label{fig:confusion_matrix}
\end{figure}

In addition to VideoMAE \cite{tong2022videomae}, we evaluated ViVit \cite{arnab2021vivit} and TimeSformer \cite{bertasius2021space} as backbone architectures for our framework. 
The performance of all three models is summarized in Table \ref{tab:performance}. 
Due to VideoMAE's superior performance and lower hardware requirements, both alternative approaches were discarded, and VideoMAE was selected as the backbone for our experiments.
The proposed model achieved a balanced accuracy of 72.2\% on the validation set. 
This performance increased to 75.7\% when multiple samples per video were aggregated using a simple majority voting approach. 
Figure \ref{fig:confusion_matrix} presents the confusion matrix for the validation dataset on the level of individual clips.

\begin{figure}[!b]
  \centering
  \includegraphics[width=1.\textwidth,trim=.3cm 0cm .2cm 0cm,clip]{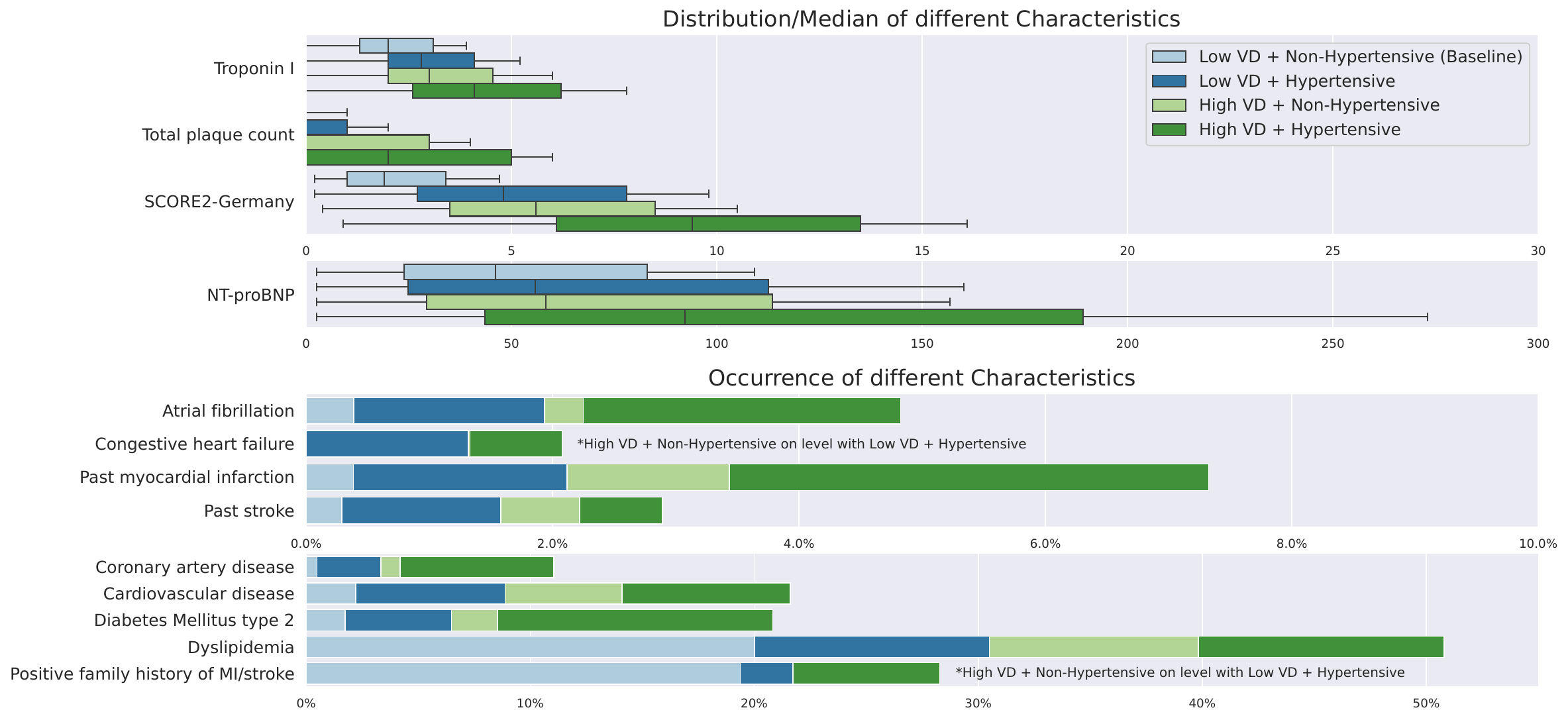}
  \caption{Statistical comparison of the classification results on the validation data with respect to different clinical parameters. For all variables the individuals classified as “high visual arterial damage” (high VD) show a significant higher likelihood or median value than individuals with low visual damage (low VD). Individuals with high visual damage and a positive hypertension diagnosis exhibit the worst cardiovascular health condition (in terms of the presented indicators).}
  \label{fig:comp}
\end{figure}

\subsection{Comparing classification results}
A comparison of the classification results on the validation data shows that individuals classified as high VD, regardless of their hypertension diagnosis show a drastically worse cardiovascular health condition (in terms of the presented indicators) than individuals rated as low VD.
Individuals with a positive hypertension diagnosis and a detected high VD presented with the worst cardiovascular health condition, followed by individuals with high VD but no hypertension diagnosis (Figure \ref{fig:comp}). 

However, using the non-hypertensive individuals with low VD as baseline, we compare our results for (1) comorbidities, (2) traditional risk markers, (3) laboratory parameters, (4) other metrics and (5) future incidents.

\subsubsection{Comorbidities:}
Individuals classified as non-hypertensive with high VD exhibit a 1.9-fold higher likelihood of dyslipidemia and a 4.9-fold higher likelihood of diabetes mellitus type 2 compared to the baseline of non-hypertensive individuals with low VD. 
However, the likelihood of a family history of MI or stroke is comparable among non-hypertensive individuals, regardless of whether they are classified as having high or low VD.

\subsubsection{Traditional Risk-Markers:}
Individuals classified as non-hypertensive with high VD demonstrate a 6.4-fold higher likelihood of CVD, an 8.7-fold higher likelihood of coronary artery disease, an 8.9-fold higher likelihood of a history of MI, and a 7.7-fold higher likelihood of a history of stroke compared to the baseline group. 
Atrial fibrillation shows an increased likelihood of 5.8-fold, while congestive heart failure did not occur once in the baseline group, while 1.3\% of all individuals in the non-hypertensive with high VD group had congestive heart failure.

\subsubsection{Laboratory Parameters:}
Individuals classified as non-hypertensive with high VD exhibit a 50.0\% higher Troponin I and a 26.4\% higher in NT-proBNP median-measure compared to the baseline of non-hypertensive individuals with low VD. 

\subsubsection{Other:}
Individuals classified as non-hypertensive with high VD show a 5.1-fold increased average total plaque count and a 2.4-fold increased SCORE2 median compared to the baseline of non-hypertensive individuals with low VD.

\subsubsection{Future Incidents:}
A total of 974 adverse cardiovascular events were observed within a 5-year interval and 1,229 adverse cardiovascular events within a 10-year interval in GHS. 
Of these, 200 incidents within a 5-year interval and 255 incidents within a 10-year interval occurred in the validation dataset.

\begin{figure}[!b]
  \centering
  \includegraphics[scale=.45,trim=1cm 0cm 0cm 0cm]{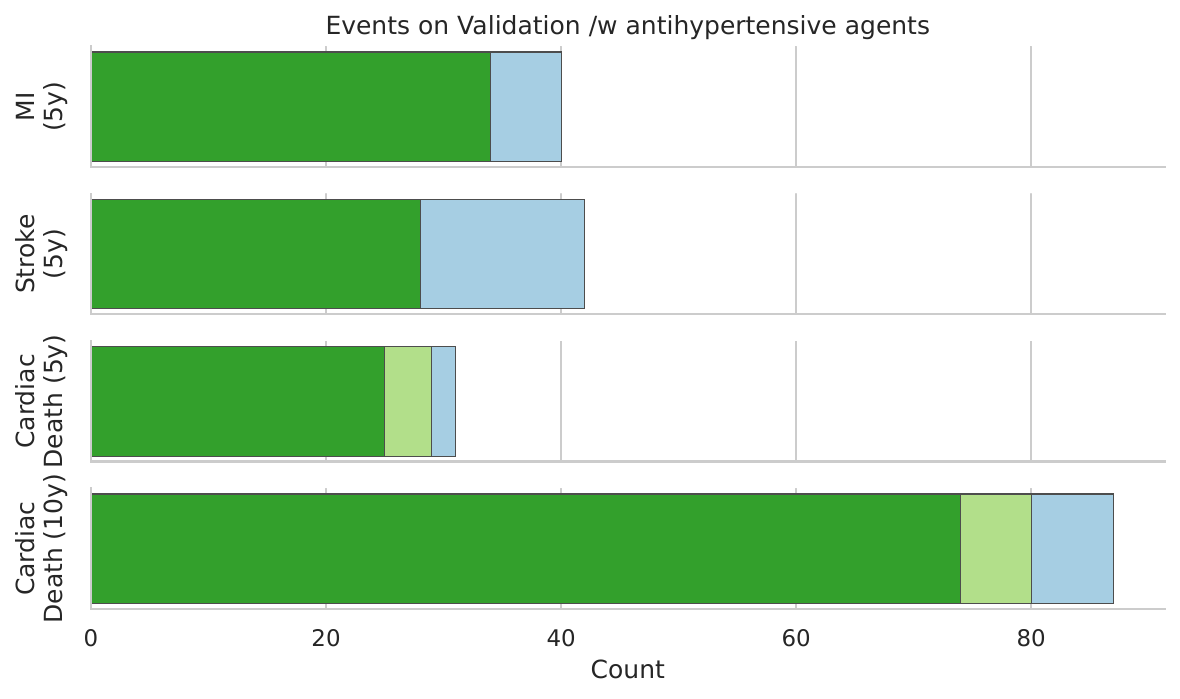}
  \includegraphics[scale=.45,trim=1cm 0cm 0cm 0cm]{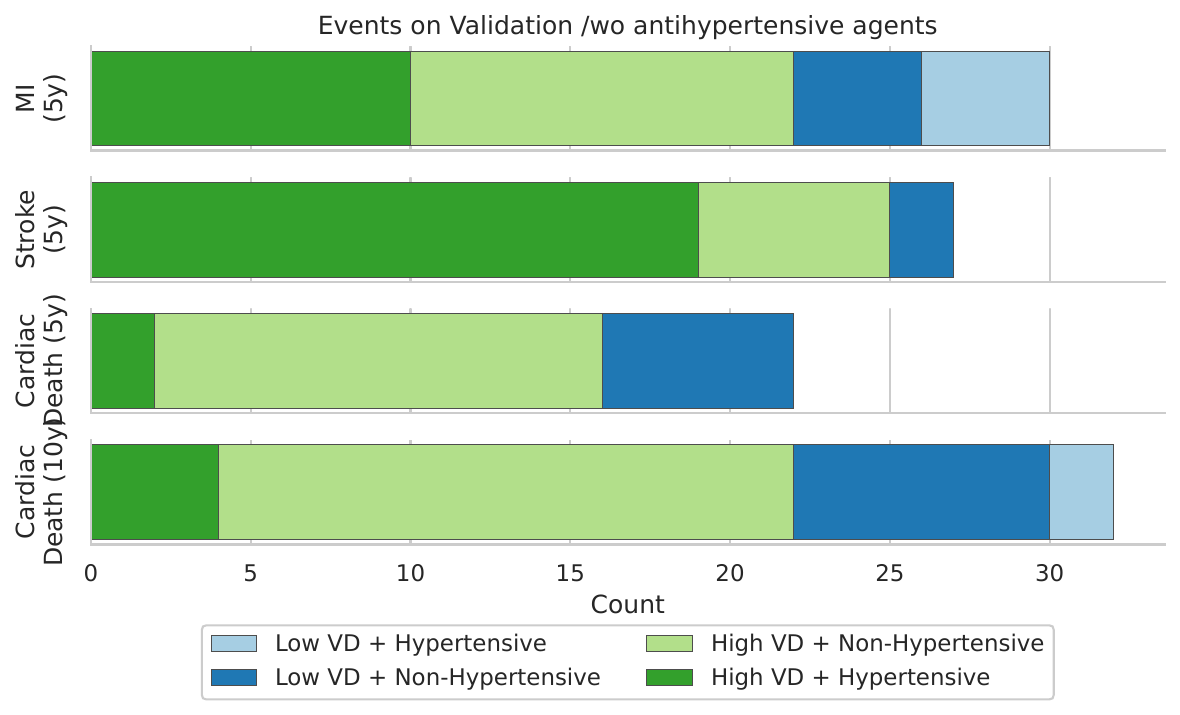}
  \caption{The total event counts in the validation dataset with respect to the model’s classifications and the use of antihypertensive agents. The upper section considers individuals taking antihypertensive agents, while the lower section focuses on those not using such medications. In both groups, the role of hypertension as a major risk factor for various cardiovascular events is emphasized by the observed incidents within the GHS cohort. Among individuals taking antihypertensive agents, the majority of events are observed in those with a positive diagnosis of hypertension and high visual arterial damage. However, in cases of untreated hypertension, individuals without hypertension but exhibiting high visual damage account for a significant proportion of all events, regardless of the type of event.}
  \label{fig:future_events}
\end{figure}

Figure \ref{fig:future_events} illustrates the distribution of individuals across various incident categories, with the data further stratified by the use of antihypertensive medications. 
Comparative analysis of the different groups indicates that individuals with high VD constitute 81.3\% of all events in the absence of antihypertensive treatment and 79.8\% of all events among those receiving treatment. 
Notably, in the absence of treatment, individuals with high VD represent the majority of incidents. 
Within this cohort, the subgroup of non-hypertensive individuals with high VD demonstrates a markedly elevated incidence rate compared to other groups.

\section{DISCUSSION}
\subsection{Classification Performance}
The model demonstrated a balanced accuracy of $75.7\%$ on the validation set (2,847 individuals) when combining multiple sampled clips per video. 
It is important to emphasize that the goal is not to achieve perfect detection of the actual blood pressure condition. 
Given that direct measurement of hypertension, as defined by actual blood pressure, is not feasible through ultrasound without Doppler imaging, the model must rely on identifying visual features that exhibit the strongest correlation with hypertension. 
Among these features, arterial damage—particularly that associated with untreated hypertension \cite{cushman2003burden}—provides a clear and reliable basis for the model to generate accurate predictions.

\begin{figure}[!b]
  \centering
  \includegraphics[scale=.55,trim=0cm 0cm 0cm 0cm,clip]{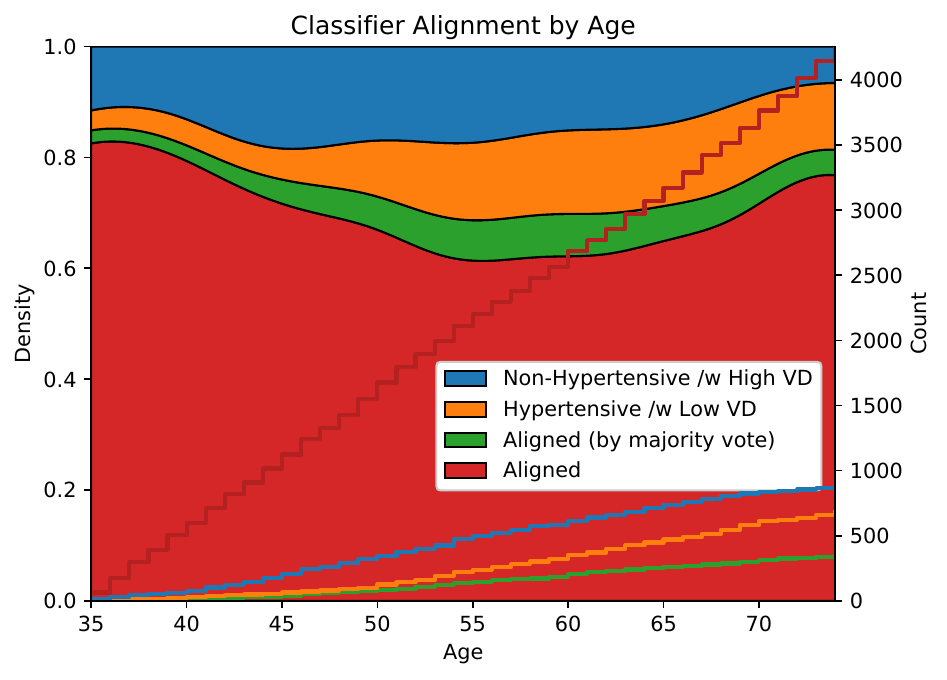}
  \caption{The background displays a kernel density estimation which illustrates the proportion of the classification alignment on a video level in relation to the individuals' age. The term 'aligned' in this context refers to the alignment of the classifier prediction of high visual damage with the clinical diagnosis of hypertension, and conversely, the alignment of a detected low visual damage with a non-hypertensive diagnosis.
  The plot in the foreground shows the count of the videos with respect to the individuals' age.}
  \label{fig:kde_age}
\end{figure}

An analysis of the classifier's performance relative to the individuals' age, as illustrated in Figure \ref{fig:kde_age}, reveals a performance dip among individuals aged 52 to 62. 
The underlying reasons for this dip warrant further investigation. 
Notably, majority voting, achieved by utilizing multiple samples per video, appears to play a critical role in enhancing model performance during this age range.
This finding prompts critical questions regarding the mechanisms through which majority voting facilitates improved classification accuracy.

It is plausible that the model's predictions are based on subtle nuances within the video data that may not always be discernible. 
These nuances may only be visible in one or even only part of a specific phase of the cardiac cycle. 
However, due to the uniform-random extraction of video clips, these critical features may not be consistently present across all clips.

Despite these challenges, the model behaves in a manner consistent with our expectations. 
Specifically, it exhibits a low rate of non-hypertensive with high VD and a higher rate of hypertensive with low VD among younger individuals. 
Conversely, in older individuals, hypertensive with low VD decrease while non-hypertensive with high VD increase. 
This pattern aligns with the anticipated progression of arterial health deterioration associated with ageing \cite{jani2006ageing,mcgrath1998age}.

\subsection{Comparison with Clinical Parameters}
An analysis of the presented clinical parameters reveals that individuals with high level of VD, irrespective of their hypertensive status, show the highest expression levels across all assessed parameters.
Notably, non-hypertensive individuals with low VD are associated with the lowest cardiovascular risk.
Furthermore, hypertensive individuals with low VD display a closer alignment in presented clinical characteristics to non-hypertensive individuals with low VD than to those with high VD.
Moreover, the SCORE2 risk prognosis aligns with our predictions of VD, further supporting the reliability of our proposed approach.
In particular, in combination of high VD and hypertension, the individuals show a drastically increased SCORE2 risk prediction.

In general, both non-hypertensive individuals exhibiting high VD and hypertensive individuals with low VD appear to represent edge cases. 
Hypertensive individuals with low VD are associated with relatively favourable arterial health, characterized by the absence of visible negative risk indicators. 
This suggests that VD may either not yet be present or remains below the detection threshold implicitly employed by the model to distinguish between the two classes.
Conversely, non-hypertensive individuals with high VD are typically linked to poor arterial health as evidenced by the presence of significant negative visual features. 
This observation suggests two potential scenarios: either hypertension was undiagnosed at the time of the baseline assessment within the GHS cohort, with the VD already exceeding the model’s implicit detection threshold, or these individuals are clinically non-hypertensive based on blood pressure measurements yet still display VD. 
Importantly, such VD may arise from factors unrelated to hypertension.

\subsection{Comparison of Future Incidents}
A similar pattern is evident in the future incidents reported by the GHS, mirroring the behaviour observed in clinical parameters. 
Hypertension, as a key co-occurring factor associated with future major cardiovascular events \cite{eto2005impact,alderman1999diabetes}, is reflected in the model’s performance, particularly in its ability to align hypertension with a detected high level of VD. 
However, a significant proportion of major cardiovascular events within the GHS cohort without antihypertensive drug usage occur among non-hypertensive individuals who exhibit high levels of VD.

Importantly, the model appears to extract ultrasound-derived features that demonstrate a strong association with general cardiac mortality and MI.
This highlights the utility of predicted VD generated by the machine learning framework and underscores its relevance to assessing general cardiovascular health at the individual patient level.




\subsection{Visual Arterial Damage}
As direct measurement of hypertension in terms of actual blood pressure is not feasible using ultrasound (without Doppler), it is assumed that the model relies on proxy indicators. 
The model is trained exclusively to classify hypertension by extracting features in an unsupervised manner, with the objective of minimizing the binary cross-entropy loss. 
A notable limitation of this approach is that during training, the model is solely reliant on hypertension labels. 
These labels may be positive even in cases where no VD is present, and may be negative despite the presence of ultrasound-detectable VD.

While it cannot be conclusively demonstrated that the features extracted by the model have a causal relationship with major cardiovascular events or represent the co-occurrence of diverse cardiovascular risks, it is evident that these features exhibit significant correlations with such phenomena.

The observed discrepancies between the independent indicators of hypertension and VD suggest that the identified VD marker should not serve as a replacement for hypertension but rather as a complementary digital risk indicator. 
This digital marker encapsulates a range of effects observable in the carotid artery.

The VD marker reveals correlations with a broad range of clinical and laboratory parameters that are well-established and widely recognised as valuable for cardiovascular risk assessment.
Importantly, it remains unaffected by phenomena such as white-coat hypertension, wherein an individual's blood pressure is elevated in clinical settings but normal in other environments, such as at home \cite{franklin2013white}.

Furthermore, carotid sonography is a fast, accessible, and cost-effective diagnostic tool. 
Its implementation is feasible across diverse settings and does not require sophisticated medical infrastructure, making it a valuable addition to cardiovascular risk evaluation protocols.

\section{CONCLUSION}
In this study, we utilized over 31,000 carotid sonography videos from the Gutenberg Health Study to train a deep learning model for classifying individuals based on the presence of high or low visual arterial damage (VD) using hypertension as a reference. 
Additionally, we demonstrated that in the absence of a direct method to measure blood pressure via ultrasound, the model effectively relies on proxy features for classification.

Through an analysis of the model's predictions in conjunction with clinical parameters, we validated the relevance of the proxy features extracted. 
By comparing the predicted levels of VD with individuals' hypertension diagnoses, we established that the extracted features align with clinical expectations across a range of parameters, including traditional risk markers, comorbidities, and laboratory findings. 
These results indicate that, from a clinical perspective, cardiovascular risk is significantly elevated when our machine learning pipeline predicts high VD, even in the absence of a formal hypertension diagnosis.

Moreover, our analysis of predicted levels of VD in relation to future major cardiovascular events revealed that individuals classified as having high VD account for the majority of such events, even if they lacked a hypertension diagnosis. 
This pattern was particularly pronounced in hypertensive individuals with high predicted VD.

Encouraged by these findings, we advocate for the development of novel, individualized risk markers leveraging machine learning approaches. 
Such markers have the potential to enhance accessibility to comprehensive cardiovascular risk assessments, mitigate the confounding effects of conditions like white-coat hypertension, and improve cardiovascular prevention strategies for the broader population.

\subsection{Future Research}
Our hypothesis regarding the extracted features suggests that the model primarily focuses not on blood pressure directly but on indicators of damage to the vessels, vessel walls, and surrounding tissue. 
However, as non-experts in the medical domain, we cannot draw definitive conclusions about specific causal medical dependencies associated with the extracted features. 
Nonetheless, our findings demonstrate a strong correlation between the extracted features and clinical parameters as well as future cardiovascular incidents.

Consequently, we present our findings to the research community to stimulate further investigation into the potential of AI-based ultrasound assessments and AI-driven CVD risk assessments. 
For future research, we aim to conduct a more detailed analysis of various clinical variables to evaluate their utility as potential proxies for risk prediction.

Building on the promising results of this study, our overarching goal is to develop an innovative, patient-centered approach to risk assessment. 
In addition, we intend to explore the integration of multiple proxy classifiers as “weak classifiers” combining their outputs to predict a comprehensive risk score for an individual's cardiovascular risk and future major events. 
Furthermore, we plan to apply explainable AI (XAI) methodologies to elucidate potential medical explanations for our findings, thereby narrowing the gap between model predictions and clinical interpretability.

\begin{credits}
\subsubsection{\ackname} This work was part of the cluster for atherothrombosis and individualized medicine (curATime), funded by the German Federal Ministry of Education and Research (03ZU1202KA).

\subsubsection{\discintname}
The authors have no competing interests to declare that are relevant to the content of this article.
\end{credits}

%
%
%
%

\bibliographystyle{splncs04} 
\bibliography{refs} 

\end{document}